\documentclass[letterpaper, 10 pt, journal, twoside]{ieeetran}

\author{Shiladitya Dutta$^{1}$, Aayush Gupta$^{1}$, Varun Saran$^{1}$, and Avideh Zakhor$^{1}$%
\thanks{Manuscript received: August, 10, 2025; Revised October, 27, 2025; Accepted November, 19, 2025.}
\thanks{This paper was recommended for publication by Editor Soon-Jo Chung upon evaluation of the Associate Editor and Reviewers' comments.} 
\thanks{$^{1}$All authors are with the College of Engineering, Department of Electrical Engineering and Computer Science, University of California Berkeley
        {\tt\footnotesize shiladitya\_dutta@berkeley.edu}}%
\thanks{Digital Object Identifier (DOI): see top of this page.}
}

\usepackage{graphics} 
\usepackage{epsfig} 
\usepackage{mathptmx} 
\usepackage{times} 
\usepackage{amsmath} 
\usepackage{amssymb}  
\usepackage{booktabs}
\usepackage{array}
\usepackage{multirow}

\title{
Vision-Guided Outdoor Flight and Obstacle Evasion \\ via Reinforcement Learning
}

\begin{document}

\maketitle


\begin{abstract}
Although quadcopters boast impressive traversal capabilities enabled by their omnidirectional maneuverability, the need for continuous pilot control in complex environments impedes their application in GNSS and telemetry-denied scenarios. To this end, we propose a novel sensorimotor policy that uses stereo-vision depth and visual-inertial odometry (VIO) to autonomously navigate through obstacles in an unknown environment to reach a goal point. The policy is comprised of a pre-trained autoencoder as the perception head followed by a planning and control LSTM network which outputs velocity commands that can be followed by an off-the-shelf commercial drone. We leverage reinforcement and privileged learning paradigms to train the policy in simulation through a two-stage process: 1) initial training with optimal trajectories generated by a global motion planner acting as a supervisory backbone, 2) further fine-tuning in a curriculum environment. To bridge the sim-to-real gap, we employ domain randomization and reward shaping to create a policy that is both robust to noise and domain shift. In outdoor experiments, our approach achieves successful zero-shot transfer to both obstacle environments and a drone platform that were never encountered during training. 
\end{abstract}


\section{INTRODUCTION}

\IEEEPARstart{R}{emote} controlled quadcopters are popular within commercial and enterprise markets for their unparalleled mobility in a small form factor. However, their control process presents key drawbacks including bounds to where they can traverse due to wireless connection limitations and the need for constant operator attention. While autopilot for preset simple paths is a common feature, in obstacle-rich, dynamic or unknown environments autonomous navigation still is not fully realized. This impedes their utilization in GNSS and telemetry denied applications such as sub-canopy forest flight, underground mapping, war zones, and industrial inspection.

As such, an open area of research is autonomous navigation in unknown environments using only onboard computation and sensors, typically in the form of vision-based systems. While traditional methods decompose this task into planning, perception and control units, new end-to-end learning-based methods using privileged information are promising \cite{loquercio2021learning} \cite{song2023learning}. These involve creating an expert policy with full environmental and state data to generate optimal paths, then distilling it into a student policy using supervised learning. However, these methods use body rates or trajectories followed by a model predictive controller (MPC) to achieve agile flight which cannot be used as-is across different drone \cite{loquercio2021learning}.

This paper aims to train a policy that can be directly deployed onto a real drone using a velocity command API, without requiring drone-specific tuning.
We accomplish this by pairing reinforcement learning (RL) and privileged learning paradigms to train an end-to-end model that outputs reference velocity commands which can be followed by off-the-shelf consumer drones through built-in APIs. We take a modular approach to constructing this network where a pre-trained AutoEncoder acts as a perception head by mapping the depth image data to a low-dimensional latent space. We then freeze the perception component and train an LSTM planning/control network first with optimal trajectories as a supervisory backbone for the reward function before further fine-tuning on a curriculum environment. To cross the Sim2Real gap, we employ domain randomization and policy smoothing to ensure successful transfer on varying environments and platforms. Domain randomization teaches the policy to be robust to sensor noise and trains the LSTM to account for differing system dynamics such as latency, inertia, etc. across time-steps \cite{gronauer2023comparing}. Meanwhile, reward shaping prevents large action fluctuations that would degrade the drone's stability and controllability.

\begin{figure}
    \centering
    \includegraphics[width = 0.38\textwidth]{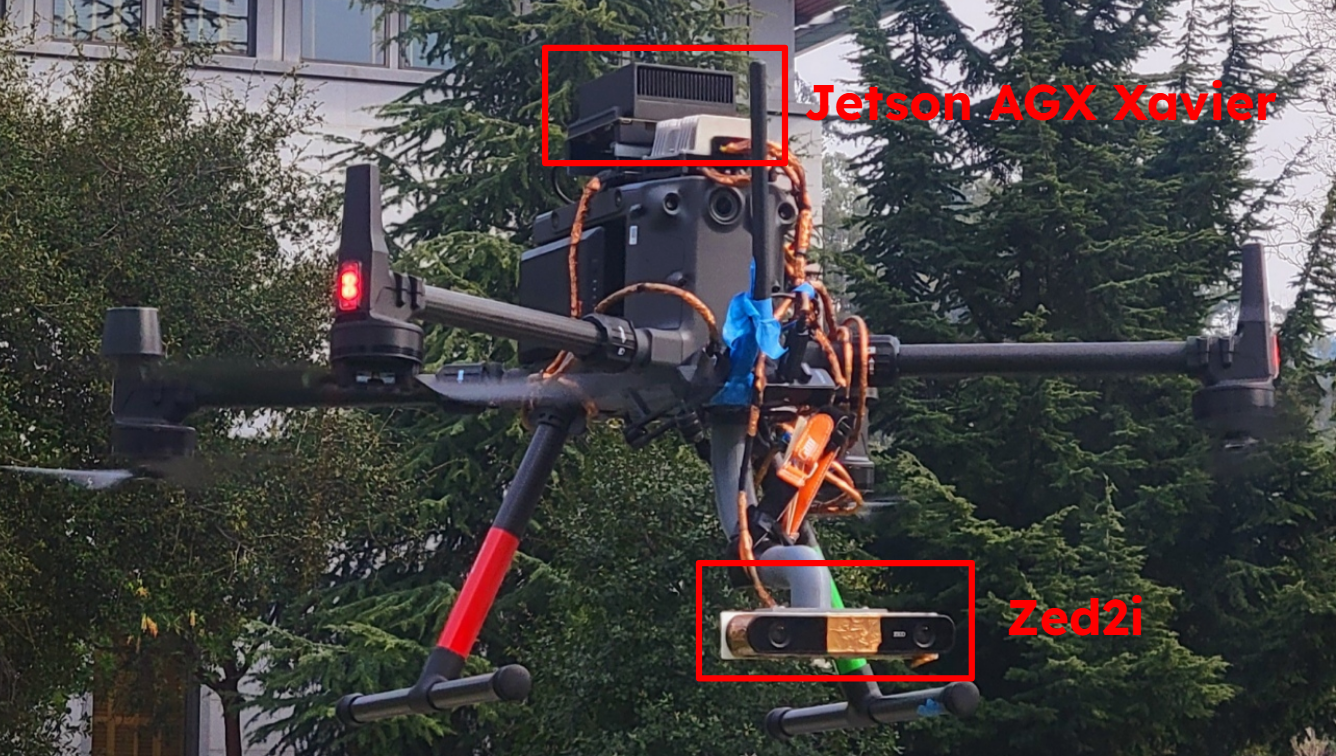}
    \caption{Actual testbed is a DJI M300 with an attached Zed2i sensor for depth estimation \& VIO and a Jetson AGX Xavier for compute.}
    \label{fig:hardware}
    \vspace{-.2in}
\end{figure}
\begin{figure*}[ht]
    \centering
    \includegraphics[width = 0.8\textwidth]{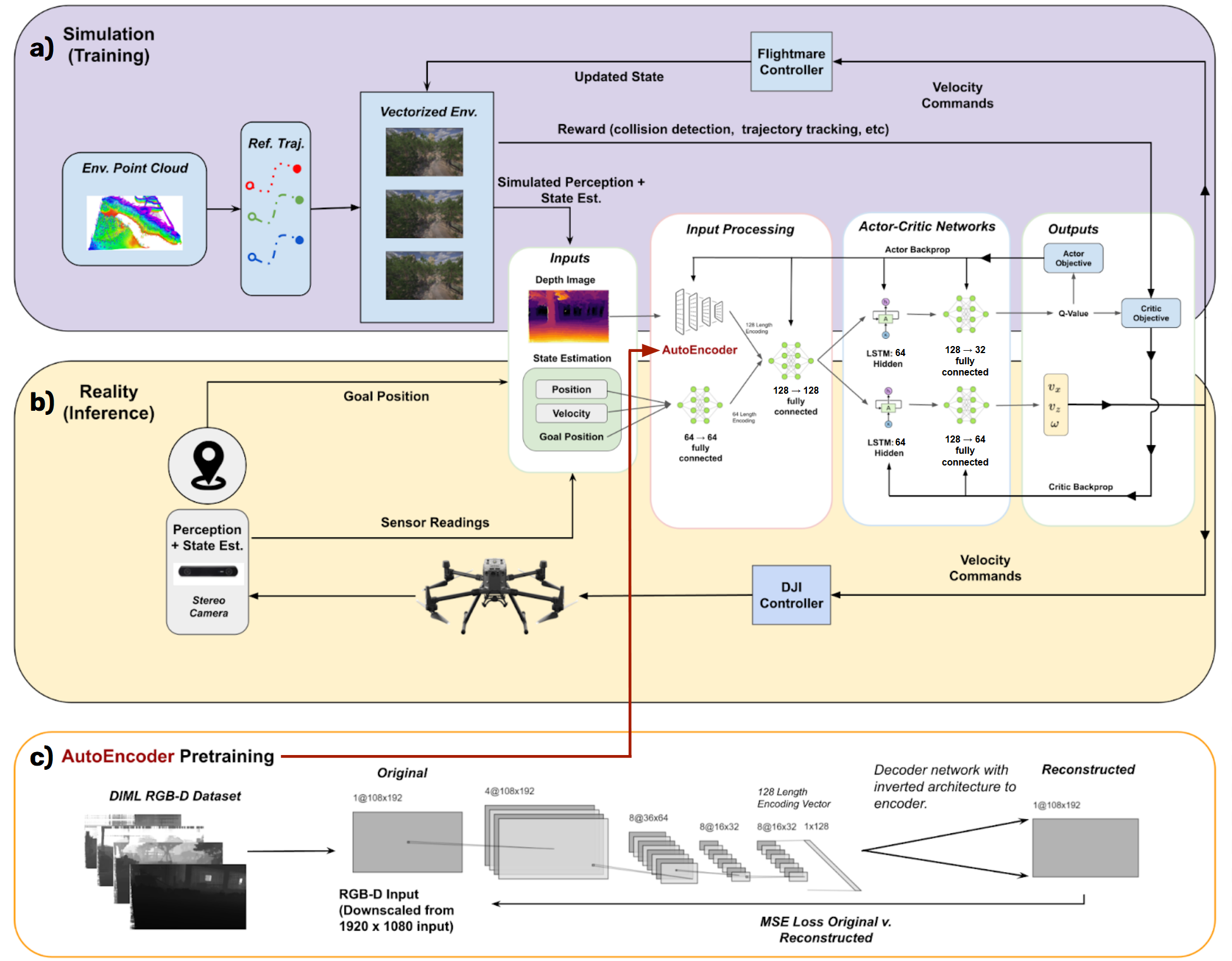}
    \caption{Overview of System. To train the input processing and actor-critic networks in (a), we use PPO for optimization and Flightmare for simulation. When deploying this policy in reality with (b), we interface with the drone using the DJI Controller and collect state estimates + depth image using the stereo RGB-D camera. In (c), we train an auto-encoder to process depth images which is used as the perception head of the policy during (a) and (b).}
    \label{fig:overview}
    \vspace{-.2in}
\end{figure*}

The result is a system that can navigate through obstacles to a goal in an outdoor environment using stereo-vision depth and visual inertial odometry (VIO). With real-world experiments covering a total of 650m in field trials, the policy overcomes Sim2Real gaps such as scenery changes, sensor noise, background clutter, and crosswinds to demonstrate successful and accurate collision-free navigation in new  environments. Furthermore, it achieves this on an off-the-shelf drone testbed --depicted in Figure \ref{fig:hardware}-- with dynamics and characteristics substantially different from those encountered during training at more than 10 times the weight of the simulated drones, highlighting the system's robustness.


\section{RELATED WORK}

There has been extensive research on methods for collision-free drone flight in cluttered environments. For navigation in unknown environments, a common strategy is to use a reactive planner which filters through a set of feasible trajectories based on some form of occupancy map to generate a local plan.
Much of the research in this area focuses on optimizing the trajectory search space and collision-checking processes. Examples include bitwise trajectory elimination \cite{viswanathan2020efficient} and rectangular pyramid partitioning for efficient collision checking \cite{bucki2020rectangular}. Despite these optimizations, end-to-end neural policies remain architecturally more efficient, as their inference is significantly simpler and more conducive to hardware acceleration than tasks such as map fusion. In addition, many classical approaches --when integrated together-- react adversely to compounding errors in state estimation, system latency, and sensor noise. As such, learning-based approaches replacing some or all of this stack have become a focal point of research in recent years \cite{hanover2024autonomous}. In particular, end-to-end sensorimotor policies that directly map vision to action have proven to be effective in agile high-speed scenarios \cite{fu2023learning}.

The architecture of these end-to-end networks can be similarly broken up into a perception section and a planning/control section. For the perception network, common practices are to employ convolution layers or an image autoencoder pretrained on a separate image dataset. Other approaches use neural monocular depth estimation to eliminate the need for stereo cameras \cite{zheng2024monocular} and to train the autoencoder with collision meshes so that the perception network outputs can be collision-aware \cite{kulkarni2024reinforcement}. For control, many monocular approaches either have the perception network directly predict a steering angle and collision probability \cite{Loquercio2018DroNetLT} or follow a state machine \cite{McGuire2017Efficient}. For end-to-end sensorimotor approaches, while fully connected networks are commonly used for the planning/control portion, some studies have proposed using Long Short-Term Memory (LSTM) \cite{gronauer2023comparing} or attention layers \cite{singla2019memory} for their ability to implicitly compensate for sim2real factors and remembering partially observed environments over multiple time-steps. A diverse range of action outputs have been explored from generating local trajectories in a receding horizon \cite{loquercio2021learning} to directly outputting motor commands \cite{Ferede2023EndtoEndRL}. 

To train these networks, either reinforcement learning or imitation learning is typically used. Deep RL methods employ a drone dynamics simulator such as AerialGym \cite{kulkarni2023aerialgymisaac} to train a policy, though some works also integrate real-world data as part of the training loop to account for the sim2real gap in flight dynamics \cite{kang2019generalization}. With imitation learning, a privileged expert provides a supervisory signal to a learned student policy, usually employing a distillation approach where the distance between the student and teacher output is optimized. For the expert, some works use traditional global planning methods \cite{loquercio2021learning} while others use a Deep RL agent trained with privileged information and a perception-aware reward \cite{song2023learning}. 
However, our approach fundamentally differs in that these works employ imitation learning to train policies specifically tuned to a drone's flight dynamics, whereas we use reinforcement learning to train a policy that can bridge the sim-to-real gap to a drone with significantly different characteristics to those encountered during training.
We will compare our approach to \cite{loquercio2021learning} and \cite{song2023learning} with more detail in Section \ref{sec:conclusion}.

Previous works have investigated combining expert data into policy learning, thus laying the groundwork for our approach.
The most notable of these is DAgger \cite{ross2011reduction} which proposes an iterative imitation learning algorithm that mitigates compounding errors by training a policy on a dataset that is aggregated with expert-labeled actions from previously visited states.
There is also \cite{zhai2022computational} which empirically and theoretically compares the benefits of sparse terminal rewards to intermediate subgoal rewards in goal-reaching policy learning.
Lastly, \cite{zhang2024learningbasedquadcoptercontrollerextreme} trains a low-level controller that can adapt to quadcopters across varying sizes by combining privileged learning from an expert model-based controller with reinforcement learning.

\section{METHODOLOGY}
Our approach --illustrated in Figure \ref{fig:overview}-- trains a neural network policy to autonomously navigate a drone to a target while evading obstacles. The policy processes depth images and state estimates from onboard sensors to generate velocity commands for the drone to follow. We construct this policy in two stages. We first pre-train an autoencoder shown in Figure \ref{fig:overview}c to embed depth images into a latent representation. Then, as shown in Figure \ref{fig:overview}a, we freeze the image encoder while training an LSTM policy network in simulation using PPO. We begin by generating a randomized training environment and computing optimal trajectories. We then train the policy in this environment where the optimal trajectories are a component of the reward function. Finally, we introduce a more complex curriculum environment and fine-tune the policy without optimal trajectory rewards, enhancing the policy's generalization.

\subsection{Model Architecture}
    
    {\bf State and Action Space} We define the drone's position at time $t$ as $\vec{p_t}=\{p_{t, x}, p_{t, y}, p_{t, z}\}$ for x, y, and z respectively. Likewise, the goal position is $\vec{g}=\{g_x, g_y, g_z\}$. The state vector at time $t$ is $\vec{s_t}=\{g_x - p_{t, x}, g_y - p_{t, y}, g_z - p_{t, z}, \phi_t, \psi_t, \omega_t, \Delta{\psi}, z_t\}$ where  $\phi_t$, $\psi_t$, and $\omega_t$ are current roll, pitch, and yaw rates respectively in rad/s; $\Delta{\psi}$ is the difference between heading towards goal and current heading in radians; and $z_t$ is the flattened $192\times108$ depth image. The action at time $t$ i.e. $\vec{a_t}=\{v_{t, x}, v_{t, z}, a_{t, \omega}\}$ consists of horizontal velocity $v_{t, x}$, vertical velocity $v_{t, z}$, and yaw angular velocity $a_{t, \omega}$. These velocity commands can be directly inputted into the built-in APIs of many off-the-shelf systems such as DJI or AR Parrot drones. 

    \begin{table}[]
        \centering
        \setlength\extrarowheight{5pt}
        \setlength{\tabcolsep}{3pt}
        \scriptsize
        \caption{Reward Terms. *only used with privileged learning}
        \begin{tabular}{@{}lll@{}}
            \toprule
            {\bf Term}                        & {\bf Expression} & {\bf Weight} \\ \midrule
            Survival                    & $-\lambda_1$  & $\lambda_1=10^{-3}$ \\
            Distance to Goal            & \(\displaystyle \lambda_2 (1-\frac{\|\vec{g}-\vec{p_t}\|_2}{\|\vec{g}-\vec{p_0}\|_2})\)      &  $\lambda_2=10^{-3}$   \\
            Heading Error               & $-\lambda_3h(\psi_t, \zeta)$   &    $\lambda_3=1/3000$  \\
            Z-position Error            & $-\lambda_4(g_z - p_{t,z})^2 \label{1}$  & $\lambda_4=10^{-3}$ \\
            $\omega$ Magnitude                   & $-\lambda_5a_{t, \omega}^2$   & $\lambda_5=1/25$        \\
            $v_z$ Direction                       & $\lambda_6\begin{cases} -1.0, & \text{if } v_{t,z} (g_z - p_{t,z}) < 0
                                            \\ 0.03, & \text{otherwise}\end{cases}$ & $\lambda_6=1/5000$ \\
            Velocity Towards Goal       & $\lambda_7 (v_{t,x} \cos(h(\psi_t, \zeta_t)))$   &   $\lambda_7=1/8000$    \\
            Acceleration                & $-\vec{\lambda_{8}} \cdot (\vec{a_t}-\vec{a_{t-1}})$    &  $\vec{\lambda_{8}}= \begin{bmatrix}
                                                                                        1/20000  \\
                                                                                        1/15000 \\
                                                                                        1/20000 
                                                                                    \end{bmatrix} $   \\
            Yaw Jerk                    & $-\lambda_9|(a_{t,\omega}-a_{t-1,\omega})-(a_{t-1,\omega}-a_{t-2,\omega})|$  & $\lambda_9=10^{-3}$ \\
            Obstacle Proximity          & \(\displaystyle -\lambda_{10}\text{ReLU}(1-\frac{min_{o \in O}{d(\vec{p_t}, o)}}{3})\)  &  $\lambda_{10}=1/3000$ \\
            Trajectory Proximity*       & \(\displaystyle \lambda_{11}\text{ReLU}(1-\frac{min_{\vec{r} \in R}{\|\vec{p_t}-\vec{r}\|_2}}{5}) \) & $\lambda_{11}=1/2000$ \\ \bottomrule
        \end{tabular}
        \label{table:rewards}
        \vspace{-.2in}
    \end{table}
    
    {\bf AutoEncoder Pretraining and Feature Extraction}
    We pre-train a denoising autoencoder using the process shown in Figure \ref{fig:overview}(c). The encoder portion of this network acts as a perception head for the policy by extracting a low dimensional encoding $\mathbb{R}_{128}$ from the high-dimensional depth image input $\mathbb{R}_{192\times108}$. This aids with policy function convergence by reducing the parameters that need to be trained and allows us to tune the perception module separately from the RL simulation loop. For the training dataset, we use the DIML dataset \cite{cho2021diml} which contains depth images from a range of settings e.g. building, construction, overpass, street, trail and inject Gaussian noise into the training input scaled by the dataset's per-pixel depth confidence metrics. This approach helps with robustness by training the encoder on both a wide-range of environment types and for the noise characteristics of stereo depth measurements. As visualized in the Input Processing portion of Figure \ref{fig:overview}(a), to extract features from the observations we feed the first 7 state values $s_{t,1:7}$ (distance to goal in x, y, and z; roll, pitch, and yaw rates; difference between current heading and heading to goal) through two fully-connected layers before being concatenated with the depth image encoding and passed through two more fully-connected layers. This yields a $\mathbb{R}_{256}$ feature vector which is fed into the actor-critic module of the policy. 
    
    {\bf Actor-Critic Networks and Policy Optimizer}
    As seen in the Actor-Critic portion of Figure \ref{fig:overview}(a), both the actor and critic networks take in the feature vector and pass it through a LSTM layer with 64-dimensional hidden and memory units. The actor/critic networks then pass the LSTM outputs through two fully-connected layers (128, 32 neurons in actor and 128, 64 neurons in critic) to obtain the action $\pi(s_t)$ and Q-value $Q^\pi(s_t, a_t)$. Note, that the policy outputs range from $[-1,1]$ and are multiplied by $v_{x,\text{max}}, v_{z,\text{max}}, \omega_\text{max}$ to obtain the action $a_t = \{v_{t,x}, v_{t,z}, \omega_t\}$ where $v_{x,\text{max}}$ is the maximum horizontal speed, $v_{z,\text{max}}$ is the maximum vertical speed and $\omega_\text{max}$ is the maximum angular speed of the drone. During training, we use Proximal Policy Optimization (PPO2) \cite{schulman2017proximal} to optimize the actor-critic and state mixing networks. 


    {\bf Reward Function}
    The reward for a state transition $R(s_{t+1}, s_{t}, a_{t})$ is calculated by adding the reward terms using the definition shown in Table \ref{table:rewards}. The parameters $\lambda_{1...11}\geq0$ are empirically chosen weights. Intuitively, these rewards terms reflect either proximity to a desired state or whether the agent is approaching a desired state. We define the beginning position of the drone as $\vec{p}_0$ and the desired heading of the drone as $\zeta = \arctan((g_x-p_{t,x})/(g_y-p_{t,y}))$. We use $o \in O$ to represent the set of all obstacles in the environment and $\vec{r} \in R$ with $\vec{r} \in \mathbb{R}^3$ to represent the set of 3D (x, y, z) positions of points on the optimal trajectory. We define function $d(\vec{p}, o)$ to be the distance between an obstacle $o$ and position $\vec{p}$. We define $h(a_1, a_2)$ to represent the difference between two yaw angles $a_1$ and $a_2$ when accounting for phase unwrapping. 
    There are four terminal rewards/states: reached goal (2.0), out of bounds (0.16), crashed (0.08), and timeout (-1.0). Reached goal is triggered if the drone reaches within 1.0m of the goal i.e. $\|\vec{g}-\vec{p_t}\|_2 \leq 1.0$. Crashed is triggered if the drone is within 1.25m of an obstacle i.e. $min_{o \in O}{d(\vec{p_t}, o)} \leq 1.25$. Timeout is if the drone is still flying for more than a maximum amount of time of 40 seconds i.e. $t_{max} \leq t, t_{max}=480$. The drone is out-of-bounds once its x or y position is 8 meters outside the range defined by the start and goal positions along that axis.

\subsection{Simulation and Environments}

    {\bf Simulation Setup} We use the open-source Flightmare platform which contains both a physics engine for quadrotor simulation and a graphics engine built on Unity that handles rendering \cite{dodgedrone2022}. The simulation cycle represented in Figure \ref{fig:overview} is as follows: the current drone state is used by the Unity module to render the depth image  which is fed alongside the state into the RL policy which then outputs velocity commands. The velocity commands are processed by Flightmare's built-in low-level controller to issue rotor outputs, which the dynamics modeling engine then uses to calculate the next drone state after a $\Delta{t}$ timestep. The simulated dynamics are based on a 0.752kg Kingfisher drone. The state is fed back into Unity and this loop continues at a rate of 12 iterations per simulated second ($\Delta{t}=0.085$). During training, an environment with 150 independently simulated drones is used to collect rollouts. 

\begin{figure}
        \centering
        \includegraphics[width = 0.49\textwidth]{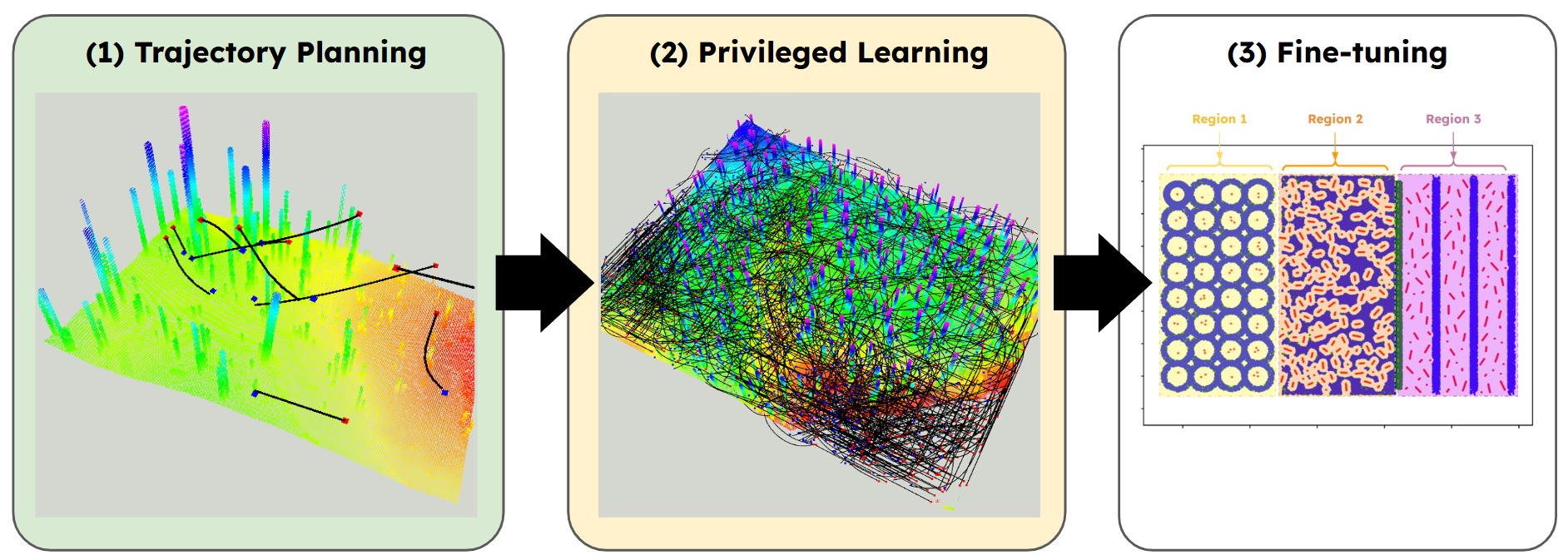}
        \caption{An overview of the 3 stages of the training pipeline: (1) generating optimal trajectories with a global planner, (2) using those trajectories for privileged learning, (3) fine-tuning in a curriculum environment.}
        \label{fig:train}
        \vspace{-.2in}
    \end{figure}

    
    {\bf Environment Generation}
    We generate two environment types: privileged learning and curriculum environments. The privileged learning environment consists of randomly placed pillars with randomized start/end locations. The curriculum environment as depicted in the fine-tuning portion of Figure \ref{fig:train} consists of three regions of increasing difficulty: 1. pillars, 2. panels, and 3. walls. Region 1 is a grid of clusters where a random number of obstacles are placed within an inner circle and start/end points are placed at varying radii in an outer circle. Region 2 consists of variable-length panels placed at random angles with start/end points being placed randomly throughout. Region 3 has variable-length walls being placed in rectangular sub-regions with start/end positions being placed on either side of each sub-region. 

    \subsection{Training}
    We train our policy using the 3 steps visualized in Figure \ref{fig:train}: trajectory planning, privileged learning, and fine-tuning.

    {\bf Optimal Trajectory Planning}
    We use Unity to convert the environment mesh into a 3D point cloud with a resolution of 0.15m. The point cloud is fed into a hierarchical motion planner \cite{liu2018search} that optimizes a dynamically feasible trajectory for time-to-goal and squared jerk for a given start/goal pair. 
    
    {\bf Privileged Learning with Trajectories}
    We teach the policy the basics of optimal time flight behavior in a simple environment of spaced-out pillars. This is aided via privileged learning whereby if the drone's position at each time step is near the optimal trajectory then it is given a reward.  We are not trying to emulate the trajectories exactly, but rather they are a guide that helps the policy converge by giving it intermediate rewards which provide a supervisory signal for planning behavior and aid convergence. This position-based approach is partially inspired by \cite{zhai2022computational} which finds that sub-goal rewards help to train goal-reaching policies.
    
    {\bf Fine-tuning in Curriculum Environment}
    After a base network is trained, we further train the policy in the more complex curriculum environment --illustrated in Figure \ref{fig:train}-- to teach the policy navigation strategies, to make it robust to environment variations, and to prevent memorization. Over the course of fine-tuning, we start off in the easiest region (1) and gradually mix in start/goal pairs from the harder regions (2/3) as training progresses. We do not use optimal trajectories during fine-tuning because the space of optimal trajectories is multi-modal (e.g. equally valid to go right or left around a single obstacle). While this effect is negligible with the simpler environments of the privileged learning step, in the more complex curriculum environment it is exacerbated since the range of near-optimal trajectories is exponential, therefore meaning that rewards for adhering to a particular trajectory is a poor signal that impedes learning.
    
\subsection{Bridging Sim2Real}

    To close the Sim2Real gap, we focus on improving the policy's noise robustness and action smoothness. The reason for the former is intuitive -- in the real-world there is noise across the state and action space: noise in the depth image, drift in the IMU/VIO measurements, randomness in latency, errors in rotor control, etc. The intuition behind striving for smoother policies is twofold. Firstly, while jerky movements may be optimal in simulation with an idealized model of flight dynamics, in the real-world they strain the rotors and degrade control authority. Secondly, jerky movement profiles do not generalize well to quadrotors with different characteristics (moments of inertia, thrust maps, body drags, etc.), thereby impeding zero-shot transfer.
    
    {\bf Reward Shaping}
    A key problem with using reference velocities as an output is that we can not enforce trajectory smoothness via methods such as interpolation. Thus, to discourage high-frequency jerky commands, we apply penalties to yaw rate magnitude, acceleration, and jerk. To discourage low-frequency oscillations which lead to wavy motion, we apply a reward if the drone is angled towards the goal and if the its trajectory matches the smooth optimal trajectory.
    
    {\bf Domain Randomization} 
    We apply a small amount of normal noise ($\pm 2\%$) to positional state measurements and a $5\times5$ Gaussian Blur with a randomized $\sigma$ in the interval [0.1, 0.7] to the depth input image. We have found empirically that Gaussian Blur degradation most closely approximates the noise encountered in real-world outdoor testing where the long range ($>$20m) and variable lighting lead to large variations in measurements near edges if the disparity between the foreground and background depth is large. For the action space, we apply normal noise ($\pm 5\%$) to the drone's simulated movement at each time step to emulate inaccuracies in control output and outdoor factors such as wind. We also varied the flight dynamics parameters such as mass and moments of inertia uniformly by $\pm 10\%$ to simulate different drone characteristics. We also add a lag factor to approximate system delays such as forward pass processing time and publish-poll time differences between ROS nodes. 
    


\section{EXPERIMENTS}
    \subsection{Simulation Evaluation}
    
    \begin{table}[]
       \centering
       \caption{Simulation outcomes across maximum speeds}
       \begin{tabular}{@{}lllll@{}}
       \toprule
       Speed ($v_{x,max}$)   & Success & Crash & Timeout & Out of Bounds \\ \midrule
       1.0 m/s &    992     &    8   &    0     &     0          \\
       2.0 m/s &    985     &   13    &    1     &      1         \\
       3.0 m/s &    989     &   10    &    1     &      0         \\
       4.0 m/s &    967     &   28    &    5     &    2           \\
       5.0 m/s &    943     &   44    &    3     &    10           \\
       6.0 m/s &    878     &   110    &    1     &    11           \\
       7.0 m/s &    780     &   265    &    2     &    53           \\
       \bottomrule
       \end{tabular}
       \label{table:sim}
       \vspace{-.2in}
    \end{table}
    
    We conduct a simulation study to analyze the policy's capabilities in a more elaborate environment than can be set up in the real world. To do so, we conduct experiments across varying maximum speeds ($v_{x, max}$) of 1.0m/s - 7.0m/s. We measure the policy's performance over 1000 runs in an obstacle course randomly generated with the same regions as the curriculum environment and start/goal points sampled uniformly across the 3 regions. The success rate shown in Table \ref{table:sim} stays relatively constant across 1.0 - 4.0m/s, however it drops off at 6.0m/s - 7.0/ms. 
    This reveals a failure case of the policy which is that it isn't able to backtrack easily. If it does head into an area that is very dense, a dead-end, or leads to out-of-bounds then it doesn't have sufficient memory to backtrack and take an alternate route. 
    This may be solved with a stronger memory architecture such as attention. \cite{singla2019memory}.

    \subsection{Simulation Ablations}
    \begin{figure}[]
        \centering
        \includegraphics[width = 0.48\textwidth]{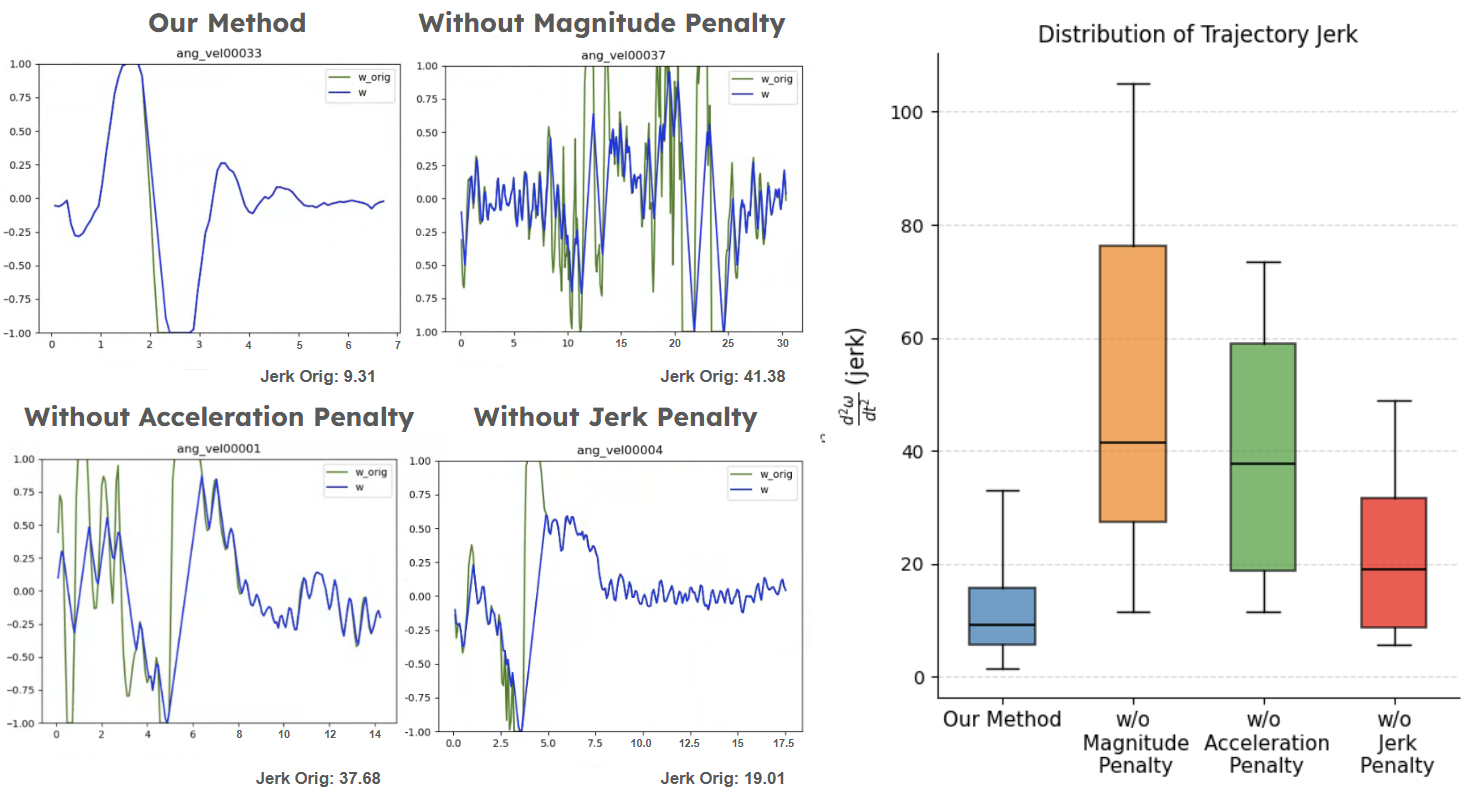}
        \caption{Ablation over reward terms. The left figures show a example plots of action commands for ablated policies in the same environment. The green line is command turn rate and the blue line is actual turn rate. The right figure shows distribution of average angular jerk over 1000 runs in the simulation environment for the ablated policies.}
        \label{fig:rew_abl}
        \vspace{-0.1in}
    \end{figure}
    
    \textbf{Ablations over Reward Terms} 
    Most of our reward terms such as distance to goal and velocity towards goal are standard for goal-reaching policy learning in that they either incentivize states near to the goal or actions that approach the goal. However, an important subset of our reward function for crossing the sim-to-real gap are those that shape trajectory characteristics: angular velocity magnitude, acceleration, and jerk penalties. Smooth trajectories are essential for reliable quadrotor control, as they enhance dynamic stability and minimize mechanical stress—hence, many motion planners aim to minimize higher-order command derivatives like jerk \cite{richter2016polynomial}. This smoothness is also important for out-of-domain generalization. By preventing erratic commands that are difficult for low-level controllers to track accurately, a low-jerk policy can generalize more effectively to platforms with varied real-world dynamics.
    To validate the effect of these reward terms, we perform an ablation study where we investigate how removing each of these reward terms from training impacts the resultant policy's behavior. As shown in Figure \ref{fig:rew_abl}, removing the magnitude penalty induces large, high-frequency oscillations in the commanded angular velocity whereas removing the acceleration penalty results in high-amplitude, low-frequency oscillations. Finally, removing the jerk penalty lends to low-amplitude, high-frequency oscillations. In addition, as seen in Figure \ref{fig:rew_abl}, removing these reward terms increases the average jerk over the trajectory of the drones in simulation.
    By penalizing the higher-order derivatives of the drone's attitude and position, we incentivize a much smoother flight profile that is more mechanically sound and robust to domain shift.

    \textbf{Ablations over Training Stages}
    \begin{figure}[]
        \centering
        \includegraphics[width = 0.48\textwidth]{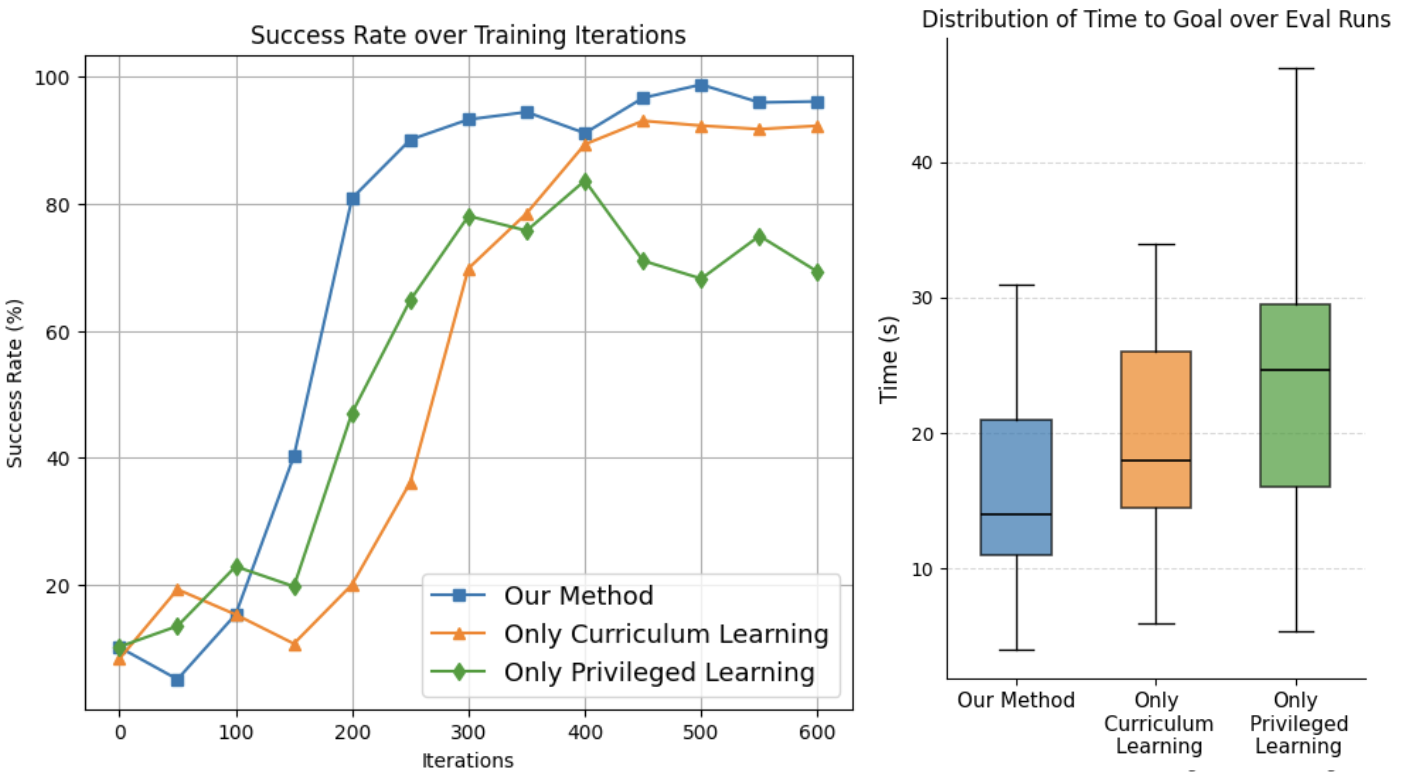}
        \caption{Simulation results in the evaluation environment at 3.0m/s for ablations over training stages. Left figure plots success rates over training iterations. The right figure shows distribution of time to goal over the same 1000 evaluation runs for the ablated policies.}
        \label{fig:train_abl}
        \vspace{-.2in}
    \end{figure}
    To validate our two-stage training paradigm, we perform an ablation study comparing our hybrid approach against policies trained with only privileged learning and only curriculum learning.
    As shown in Figure \ref{fig:train_abl}, our hybrid method achieves the highest stable success rate and lowest times-to-goal, indicating that the combination of initial guided learning followed by curriculum-based generalization is effective.
    The privileged learning policy learns quickly at the outset. However, its performance plateaus and decreases as the curriculum introduces more complex scenarios, suggesting that optimal trajectories provide a poorer reward signal in these situations. Conversely, the curriculum learning policy takes longer to learn as it relies purely on exploration. While it reaches a success rate that is closer to the hybrid method, its time to reach the goal is longer suggesting that optimal trajectories help the policy to learn jerk-efficient and time-efficient flight. These results suggest that the primary benefit of privileged trajectory learning is as a form of scaffolding, bootstrapping the policy in the early stages before its utility diminishes in more complex environments where curriculum learning proves more effective for robust generalization.
    It is plausible that with more iterations or roll-outs, the curriculum-based approach could converge to an optimal policy. Likewise, generating a large set of trajectories that approximate the multi-modal distribution of optimal paths could solve the issue of more complex environments. 
    However, unlike other works that generate multiple trajectories for local horizons, primarily for imitation learning \cite{song2023learning}), these would need to be global trajectories whose distribution has higher variance and whose generation is more computationally expensive. As such, both approaches would be significantly more computationally expensive on the training-side. 

    \subsection{Simulation Comparison}
    \begin{figure*}
        \centering
        \includegraphics[width = 0.98\textwidth]{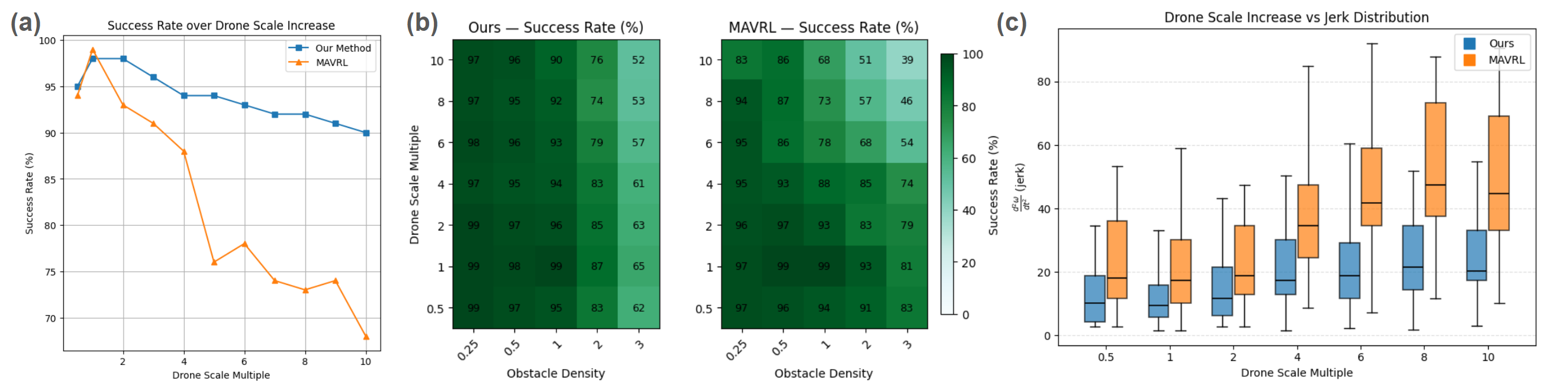}
        \caption{Charts on performance comparison to MAVRL. In (a), we compare how success rate changes as we change the scale of the drone. In (b), we display success rates over both scale multiple and obstacle density. In (c), we show the distribution of average trajectory jerk over drone scale multiple.}
        \label{fig:comparison}
        \vspace{-.1in}
    \end{figure*}
    In this section, we evaluate how our policy reacts in comparison to a baseline as the drone dynamics become more out-of-distribution (OOD) and the environments become harder to traverse. 
    For the baseline, we use MAVRL which has a similar architecture consisting of a pretrained perception network for depth imagery combined with a policy trained in simulation with RL \cite{yu2024mavrl}.
    Our approaches differ algorithmically in two primary ways: (1) MAVRL pre-trains both the encoder and LSTM whereas we only pre-train the encoder; and (2) we employ a hybrid privileged and curriculum learning strategy during RL.
    Given that both MAVRL and our method use stable-baselines as an RL framework, we directly load MAVRL's provided weights into our simulation alongside their drone dynamics which are similar to ours -- 0.72kg vs 0.752kg.

    Our experiments test generalization across two axes: drone scale and obstacle density. For drone scale, we increase the drone's weight and thrust proportionally. For obstacle density, we decrease minimum distance between obstacles, e.g. density value of 2 means halving minimum distance between obstacles.
    Our results, seen in Figures \ref{fig:comparison}a and \ref{fig:comparison}b, show how our policy is significantly more robust to out-of-distribution drone dynamics. 
    MAVRL initially demonstrates comparable or higher success rates, particularly in denser scenarios—likely due to its training in more cluttered environments and better ability to vary speed dependent on scenario. 
    However, MAVRL's performance rapidly degrades as drone scale increases. In contrast, our method maintains its success rate even as the drone dynamics deviate significantly from training. Thus, our policy is able to outperform MAVRL in 26 out of 35 tested scenarios.
    This performance is partly explained by the results shown in Figure \ref{fig:comparison}c. While MAVRL's commands are jerkier as drone dynamics become more OOD, our policy generates relatively low-jerk trajectories across the range of tested scales. These commands are easier for the low-level controller to track accurately which makes its response more consistent across drone dynamics.
    A possible explanation is that by pre-training their LSTM, MAVRL's policy develops a strong model for temporal perception but does not learn to adapt to varied drone dynamics.
    Conversely, we train our LSTM entirely within the simulation's RL loop. This process, combined with domain randomization and the reward shaping detailed by our ablation study, likely teaches the policy to generate low-jerk commands that are  more robust to shifts in the underlying flight dynamics.

    \subsection{Flight System Overview}
    All real-world tests are conducted on a DJI Matrice 300 RTK (6.3kg, 0.81m span) with additional mounted hardware as seen in Figure \ref{fig:hardware}. A Zed2i device with a stereo camera and built-in IMU is used to collect $672\times376$ resolution depth images and state estimates augmented by VIO. In the depth image, we remove any measurements over 20m to account for the Zed2i's maximum effective range and under 0.2m to account for the rotor blades being in view before downscaling to $192\times108$. The stereo depth calculations and policy are run on a Jetson AGX Xavier orchestrated with ROS. The policy interfaces with the drone using DJI's onboard SDK to send velocity commands. The full pipeline, from the depth image capture to the command being sent to the drone, runs at 12Hz. This system can ideally be deployed on other drones given the proper mounting hardware and software interface.
    \begin{figure}
        \centering
        \includegraphics[width = 0.45\textwidth]{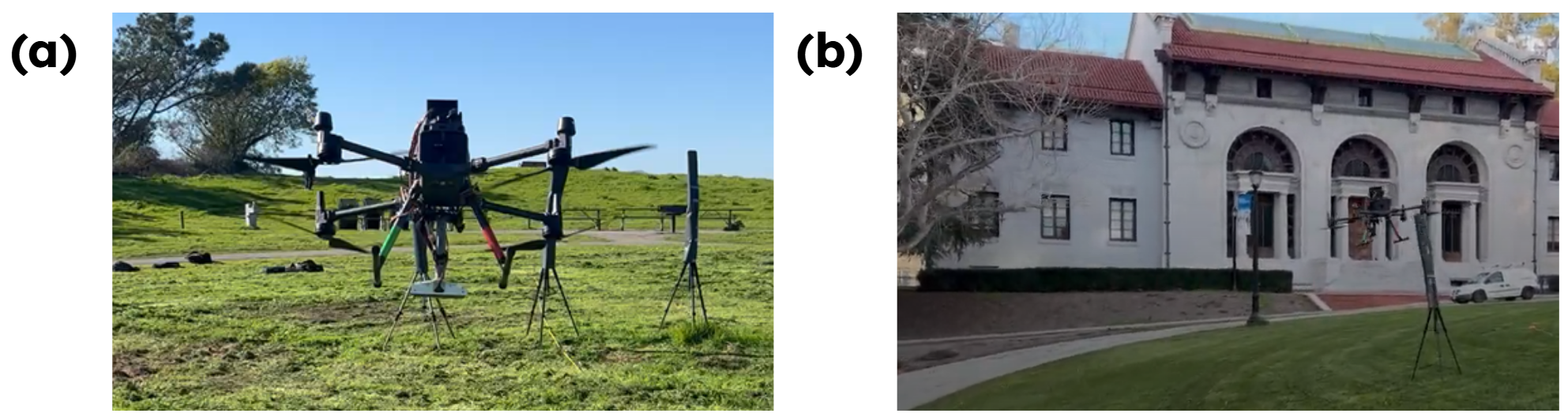}
        \caption{Pictures of the actual testing environments}
        \label{fig:realenv}
    \end{figure}
     
    \subsection{Real-World Experiments}
    \begin{figure*}
        \centering
        \includegraphics[width = 0.98\textwidth]{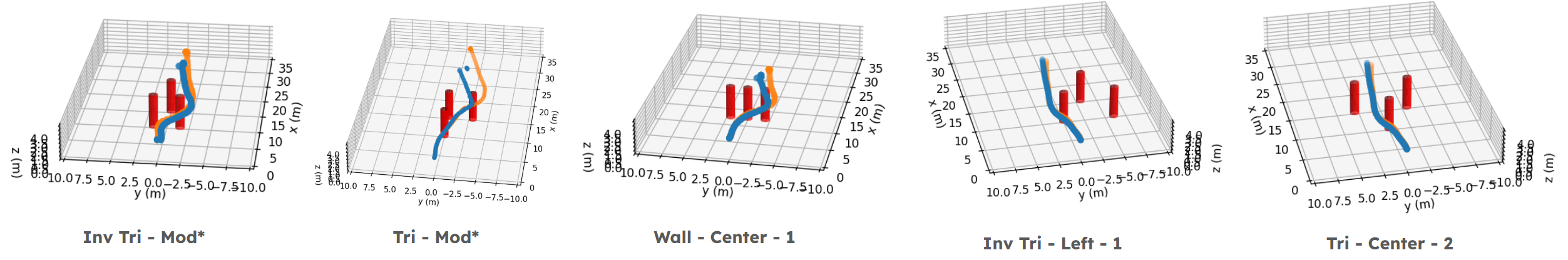}
        \caption{3D visualization of 5 runs, labeled in the form of environment - scenario - trial. Depicted are also modified environments labeled as Tri - Mod and Inv Tri - Mod which are described in the Discussion. The red pillars are the obstacles, the orange line is the GPS-measured trajectory, and the blue line is VIO device measured trajectory. Note that the blue line may seem like it is passing through obstacles, however that is due to VIO's inaccuracy.}
        \label{fig:3dplot}
        \vspace{-0.2in}
    \end{figure*}
    
    \begin{table}[]
    \centering
    \caption{VIO device positioning trials}
    \setlength{\tabcolsep}{4pt}
    {\scriptsize
    \begin{tabular}{@{}llccc|ccc|c@{}}
    \toprule
    \multirow{2}{*}{\textbf{Env}} & \multirow{2}{*}{\textbf{Scenario}}
    & \multicolumn{3}{c|}{\textbf{Trial Avg.}}
    & \multicolumn{3}{c|}{\textbf{Env Avg.}}
    & \textbf{Overall} \\
    \cmidrule(l){3-8}
    & & \multicolumn{1}{c}{s {[}m/s{]}}
      & \multicolumn{1}{c}{t {[}s{]}}
      & \multicolumn{1}{c|}{d {[}m{]}}
      & \multicolumn{1}{c}{Success}
      & \multicolumn{1}{c}{d {[}m{]}}
      & \multicolumn{1}{c|}{MP {[}\%{]}}
      & $\bar{\textrm{d}}$ {[}m{]} \\
    \midrule
    \multicolumn{1}{l|}{\multirow{3}{*}{Wall}} &
    \multicolumn{1}{l|}{Left}
      & 2.58 & 8.97 & \multicolumn{1}{l|}{2.65}
      & \multirow{3}{*}{\textbf{6/6}}
      & \multirow{3}{*}{2.65}
      & \multirow{3}{*}{\shortstack{93.53\%\\$\pm$ 3.88}}
      & \multirow{9}{*}{2.27} \\
    \multicolumn{1}{l|}{} &
    \multicolumn{1}{l|}{\textit{Center}}
      & 2.50 & 9.41 & \multicolumn{1}{l|}{2.80}
      & & & & \\
    \multicolumn{1}{l|}{} &
    \multicolumn{1}{l|}{Right}
      & 2.50 & 9.17 & \multicolumn{1}{l|}{2.50}
      & & & & \\
    \cmidrule(r){1-8}
    \multicolumn{1}{l|}{\multirow{3}{*}{Triangle}} &
    \multicolumn{1}{l|}{Left}
      & 2.56 & 9.02 & \multicolumn{1}{l|}{1.40}
      & \multirow{3}{*}{\textbf{6/6}}
      & \multirow{3}{*}{1.95}
      & \multirow{3}{*}{\shortstack{98.01\%\\$\pm$ 1.68}}
      & \\
    \multicolumn{1}{l|}{} &
    \multicolumn{1}{l|}{\textit{Center}}
      & 2.60 & 9.25 & \multicolumn{1}{l|}{2.00}
      & & & & \\
    \multicolumn{1}{l|}{} &
    \multicolumn{1}{l|}{Right}
      & 2.57 & 9.02 & \multicolumn{1}{l|}{2.45}
      & & & & \\
    \cmidrule(r){1-8}
    \multicolumn{1}{l|}{\multirow{3}{*}{\shortstack[l]{Inverted\\Triangle}}} &
    \multicolumn{1}{l|}{Left}
      & 2.67 & 8.89 & \multicolumn{1}{l|}{2.75}
      & \multirow{3}{*}{\textbf{6/6}}
      & \multirow{3}{*}{2.21}
      & \multirow{3}{*}{\shortstack{92.62\%\\$\pm$ 0.92}}
      & \\
    \multicolumn{1}{l|}{} &
    \multicolumn{1}{l|}{\textit{Center}}
      & 2.575 & 9.09 & \multicolumn{1}{l|}{1.85}
      & & & & \\
    \multicolumn{1}{l|}{} &
    \multicolumn{1}{l|}{Right}
      & 2.69 & 8.68 & \multicolumn{1}{l|}{2.05}
      & & & & \\
    \bottomrule
    \end{tabular}}
    \label{table:real}
    \vspace{-.2in}
\end{table}
    To validate that the policy successfully transfers to the real-world, we constructed 3 obstacle environments: wall, triangle, and inverted triangle. Each consists of pillars being positioned in different locations with the drone having to navigate through them to reach the goal. For each environment, we test 3 scenarios where the start and goal point are in different locations relative to the obstacles: left, center, and right. We also test two modified environments called "Tri-Mod" and "Inv Tri-Mod" in Figure \ref{fig:3dplot} where we reposition obstacles to test response to certain factors. The first is where we conceal an obstacle behind the first on the drone's expected path to see if it could quickly adapt. The second is where we reduce the obstacle from 6m to 3m to assess if the drone would avoid passages with insufficient clearance. The environments and scenarios are shown in Figure \ref{fig:3dplot} and the results are shown in Table \ref{table:real}. For the first set of experiments where we only use VIO for positioning, we performed 2 trials for each combination of scenario and environment for a total of 18 trials. We also performed a second set of experiments with GNSS positioning of 2 trials for each of the environments as shown in Table \ref{table:real2}. For VIO trials the goal was relative to the drone's starting orientation e.g. +25m forward, whereas for GNSS trials the goal was a lat/long coordinate. A run is declared successful if the drone stops within 1.0m of the goal, as measured by the positioning device used by the policy. The device for the VIO trials is the Zed2i and the device for the GNSS trials is GPS. All runs operate at a $v_{x, max}$ of 3.0 m/s. 

    \begin{table}
  \centering
  \caption{GNSS device positioning trials}
  {\scriptsize \begin{tabular}{@{}lccccc|c@{}}
    \toprule
    \textbf{Env} &
    \multicolumn{5}{c|}{\textbf{Trial Avg.}} &
    \textbf{Overall} \\
    \cmidrule(l){2-6}
    &
    \multicolumn{1}{c}{s {[}m/s{]}} &
    \multicolumn{1}{c}{t {[}s{]}} &
    \multicolumn{1}{c}{d {[}m{]}} &
    \multicolumn{1}{c}{MP {[}\%{]}} &
    \multicolumn{1}{c|}{Success} &
    $\bar{\textrm{d}}$ {[}m{]} \\
    \midrule
    \multicolumn{1}{l|}{Wall}
      & 2.63 & 8.36 & 1.05 & 97.31\% & \multicolumn{1}{l|}{2/2} & \multirow{3}{*}{0.81} \\
    \cmidrule(r){1-6}
    \multicolumn{1}{l|}{Triangle}
      & 2.51 & 8.70 & 0.65 & 97.43\% & \multicolumn{1}{l|}{2/2} & \\
    \cmidrule(r){1-6}
    \multicolumn{1}{l|}{Inverted Triangle}
      & 2.58 & 8.40 & 0.72 & 99.55\% & \multicolumn{1}{l|}{2/2} & \\
    \bottomrule
  \end{tabular}}
  \label{table:real2}
  \vspace{-.2in}
\end{table}
    
    The experiments occurred in two locations shown in Figure \ref{fig:realenv}: a field and a courtyard. The field was open with no obstacles in view whereas the courtyard had uneven terrain, a sculpture, a pond, and surrounding buildings/trees. The GNSS experiments in Table \ref{table:real2} were done in the courtyard in addition to the 6 trials for the VIO results, whose results are italicized in Table \ref{table:real}. The other 12 VIO trials were done in the field.

    Beyond the new obstacle configurations themselves, the system had to contend with multiple domain shifts to successfully cross the Sim2Real gap. The most obvious one is that the policy had never trained with the dynamics of a DJI M300 and as such had to compensate for the different flight characteristics. The environment scenery itself was also substantially different from training with the sloped terrain and clutter such as trees, a sculpture, light poles, and buildings. There were out-of-distribution environmental factors such as crosswinds reaching up to 14 kph and non-ideal lighting conditions leading to additional sensor noise. Lastly, the VIO positioning from the Zed2i was oftentimes substantially inaccurate drifting up to 3m as evidenced by GNSS measurements.

    In spite of the aforementioned Sim2Real domain gaps, the system achieved a high success rates across all 3 environments. The results for VIO positioning using a Zed2i can be seen in Figure \ref{fig:3dplot} and Table \ref{table:real} where s[m/s] is average speed across the run as measured by GNSS, t[s] is time from start to goal, d[m] is actual measured distance to goal using a tape measure, and MP [\%] is Mission Progress \cite{yu2023avoidbench}. As seen in Table \ref{table:real}, the success rate across all environments and scenarios is 100\% while the average distance to goal is 2.27m averaged over all three environments. This relatively large average distance to goal, also shown in the discrepancy between the orange and blue trajectories in Figure \ref{fig:3dplot}, are despite the fact that the VIO device believes it is within 1.0 meter of the goal 100\% of the time. This reflects the positioning inaccuracy of the chosen VIO device which limits the drone's ability to accurately reach the end goal. The drone had an average speed of 2.58m/s across the trials which is close to the $v_{x, max}$ of 3.0m/s, especially when considering the acceleration phase at the start and deceleration phase near the goal. This indicates that the policy was aiming to achieve the highest possible speed during the trials. 
    To decouple the inaccuracy of the VIO device from the performance of our policy, we conducted trials using GNSS for positioning. The results in Table \ref{table:real2} indicate a 100\% success rate and an average distance to goal of 0.81m which is 2.8 times smaller than that of the VIO experiments.
            
    
    \section{Discussion and Conclusions}
    \label{sec:conclusion}
    Our work is closely related to \cite{loquercio2021learning} and \cite{song2023learning} in that they both aim to achieve real-world vision-based flight in cluttered outdoor environments using a learning-based approach. However, this system differs from \cite{loquercio2021learning} and \cite{song2023learning} in both methodology and experimental outcomes.
    Our training approach differs from \cite{loquercio2021learning} in that they employ imitation learning to train their policy off a classical motion planner, whereas we use reinforcement learning to train our policy entirely in simulation.
    \cite{song2023learning} focuses on optimizing flight in a {\it known} environment whereas we focus on flight in unknown environments; also \cite{song2023learning} uses RL to train the expert rather than the policy itself. 
    To train the deployed policy, both \cite{loquercio2021learning} and \cite{song2023learning} use imitation learning to match the expert exactly. On the other hand, we use the relative position of the optimal vs. rollout trajectory as one of the components of the reward function during RL. 
    This distinction in how the expert provides a supervisory signal—shifting from imitation learning to reinforcement learning—fundamentally influenced the design of downstream systems, including domain randomization, reward shaping, and curriculum learning.
    
    The key takeaway is that the system is able to perform in largely out-of-distribution conditions. The policy is able to both track a goal point and avoid obstacles in spite of degraded state estimation. It is also able to generalize from the simulation dynamics of a 0.7kg drone varied only by $\pm10\%$ to a real-world drone 10 times the weight at over 7kg. These results lend credence to the idea that robust policies can use velocity commands as a layer of abstraction to interface with a broad range of drones without specific tuning. 

\bibliographystyle{plain}
\bibliography{ref.bib}

\end{document}